\documentclass[11pt]{article}

\usepackage[final]{acl}

\usepackage{times}
\usepackage{latexsym}
\usepackage[capitalise]{cleveref}

\usepackage[T1]{fontenc}

\usepackage[utf8]{inputenc}

\usepackage{microtype}

\usepackage{inconsolata}

\usepackage{graphicx}
\usepackage{paralist}
\usepackage{booktabs,enumitem}
\usepackage{multirow}
\usepackage{url}
\usepackage{hyperref}
\usepackage{tcolorbox}
\usepackage{caption}
\usepackage{subcaption}
\usepackage[dvipsnames]{xcolor}
\tcbuselibrary{listings}

\tcbset{
  boxrule=0.5pt,
  colback=white,
  colframe=black,
  arc=2pt,
  left=10pt,right=10pt,top=10pt,bottom=10pt,
  fonttitle=\bfseries\large,
  coltitle=white,
  colbacktitle=black!85!gray,
}
\usepackage{marvosym} 

\def\hmath$#1${\texorpdfstring{{\rmfamily\textit{#1}}}{#1}}

\title{Can LLMs subtract numbers?}

\author{ Mayank Jobanputra$^1$, Nils Philipp Walter$^2$, Maitrey Mehta$^3$, Blerta Veseli$^1$,\\ \textbf{Evan Parker Kelly Chapple}$^1$, \textbf{Yifan Wang}$^1$, \textbf{Sneha Chetani}$^1$, \\ \textbf{Ellie Pavlick}$^4$, \textbf{Antonio Vergari}$^5$, \textbf{Vera Demberg}$^1$ \\[1ex] $^1$Saarland University \quad $^2$CISPA Helmholtz Center for Information Security\\ $^3$University of Utah \quad $^4$Brown University \quad $^5$University of Edinburgh \\  \texttt{mayank@lst.uni-saarland.de}  }

\begin{document}
\maketitle
\begin{abstract}
We present a systematic study of subtraction in large language models (LLMs). While prior benchmarks emphasize addition and multiplication, subtraction has received comparatively little attention despite being structurally distinct as a non-commutative operation. We evaluate eight pretrained LLMs spanning four families on addition and subtraction problems. Our experiments reveal that subtraction accuracy lags behind addition by a wide margin. We find that the errors for ($a-b$) are concentrated in cases where ($a<b$). In such cases, LLMs frequently produce the correct magnitude but omit the negative sign. Probing analyses show that LLMs internally encode whether results should be negative, yet this information is often not reflected in generated outputs. 
We further test well-known techniques such as few-shot learning and instruction-tuning to see if they can improve the LLMs' performance. Our results suggest that while few-shot prompting yields modest gains, the instruction-tuned models achieve near-perfect accuracies in generating the negative sign. Together, these findings provide a clearer characterization of the limitations and recoverability of LLMs' arithmetic capabilities in subtraction.
\end{abstract}

\section{Introduction}

Large language models (LLMs) show strong performance on arithmetic tasks, particularly when aided by chain-of-thought (CoT) prompting and its variants \cite{wei2022chain, kojima2022large, imani-etal-2023-mathprompter, li-etal-2025-exposing}. However, prior work and widely used benchmarks \cite{balunovic2025mathconstruct,yang2025number,hendrycks2021math} concentrate on addition and multiplication, leaving \textit{subtraction} comparatively underexplored. 
This gap raises the question of how well LLMs handle subtraction, especially in a zero-shot setting.

Subtraction differs from addition in an important structural way as it is a non-commutative operation ($a-b \neq b-a$). This makes the order of operands decisive for subtraction.
Moreover, subtraction accentuates better positional representations to facilitate accurate borrowing.
Borrowing requires robust tracking of digit positions across potentially long sequences --- a setting where transformers trained from scratch are known to struggle~\cite{mcleish2024transformers}. These factors pose additional demands unique to subtraction, making it unclear whether LLMs can sustain their strong performance on addition~\cite{zhou2024fourier, nikankin2025arithmetic} when applied to subtraction.

In this work, we systematically evaluate subtraction across LLM families and sizes, benchmarking against addition to establish a clear performance baseline. We evaluate eight pretrained LLMs of different sizes from four model families, \texttt{Gemma-2}~\cite{team2024gemma}, \texttt{Qwen3}~\cite{yang2025qwen3}, \texttt{OLMo-2}~\cite{olmo20242}, and \texttt{Llama-3}~\cite{grattafiori2024llama}. 

We find a surprising disparity between the additive and subtractive capabilities of LLMs. Our findings reveal that pretrained LLMs struggle to generate correct answers when subtraction results are negative. Our main findings are: 

\begin{inparaenum}[a)]
    \item Under matched complexity, subtraction performance consistently lags addition in state-of-the-art open-source LLMs. In some cases, LLMs reach perfect accuracy on addition while only about half that level on subtraction. (\S~\ref{subsec:addvssub})
    \item Subtraction $a-b$ is disproportionately error-prone when $(a<b)$, with a dominant failure mode in which LLMs produce the correct magnitude but omit the negative sign. (\S~\ref{subsec:posvsneg},\ref{subsec:absolutevals})
    \item Our probing results suggest that LLMs internally encode when the result should be negative, but they often fail to surface this at decoding time. (\S~\ref{subsec:probing})
    \item Instruction tuning improves subtraction performance to levels comparable with addition in a zero-shot setting. However, adding a few-shot examples yields modest and inconsistent improvements across pretrained LLMs. (\S~\ref{sec:improvements}) This candidates subtraction as an inherently harder task than addition, and can help shed light on the different regimes (i.e. pretrained or instruction-tuned) in which LLMs learn and reason.
\end{inparaenum}

\section{Experimental Setup}
Following prior work~\cite{zhou2024fourier, nikankin2025arithmetic}, we primarily focus on \textit{single-token} numbers for our analysis. However, we also analyze \textit{multi-token} numbers in Appendix \ref{sec:oov_apdx} that show similar results.

\subsection{Data generation}
\label{subsec:data}
To evaluate LLMs on the integer subtraction task, we generate synthetic datasets using controlled settings described below.

\paragraph{Sampling operands $a$ and $b$.} 
We sample operands $a$ and $b$ uniformly from the range of numeric values that are represented as single tokens by each LLM's tokenizer (see \textit{Tokenizer Range} in Table \ref{tab:dataset_stats}). Consequently, each LLM is evaluated only on problems involving numbers within its own tokenizer range. For every sampled pair $(a, b)$, we generate arithmetic problems and categorize them based on the relationship between operands (e.g., $a > b$, $a < b$, $a = b$).
The final dataset (see \#Used Samples in Table \ref{tab:dataset_stats}) is balanced, containing an equal number of problems with $a > b$ and $a < b$. For our work, we do not make use of operands where $a = b$.

\begin{table*}[h]
\centering
\begin{tabular}{lcccccc}
\toprule
\shortstack{LLM Family} & Tokenizer Range & \shortstack{Total \\ Samples} & \shortstack{\#Samples\\ $a>b$} & \shortstack{\#Samples\\$a<b$} & \shortstack{Prompt \\ Variants} & \shortstack{\#Samples \\ Used} \\
\midrule
Qwen3    & [0, 9]    & 100        & 45       & 45       & 5 & 450 \\
Gemma-2  & [0, 9]    & 100        & 45       & 45       & 5 & 450 \\
Llama-3  & [0, 999]  & 1,000,000 & 499,500  & 499,500  & 5 & 10,000 \\
OLMo-2   & [0, 999]  & 1,000,000 & 499,500  & 499,500  & 5 & 10,000 \\
\bottomrule
\end{tabular}
\caption{
Dataset statistics for subtraction experiments across different LLM families. 
\textit{Tokenizer Range} indicates the continuous range of numeric values that are single tokens for each LLM.
\textit{Total Samples} reflects the total number of possible subtraction problems.
For LLMs where \textit{Total Samples} exceeds 2000, we uniformly sample 2000 problems from the full set. We then apply our prompt variants to each sampled problem and obtain all the samples for inference (i.e. \textit{\#Samples Used}).
}
\label{tab:dataset_stats}
\end{table*}

\paragraph{Prompt Variants.}  
To ensure the generalizability of our results and minimize the risk of spurious correlations, we use five different prompt formats, ranging from minimal equation style input to verbose template style input. We provide the exact prompt templates in the Appendix \ref{sec:prompt_apdx}.

\paragraph{Zero-shot vs.\ n-shot.}
\emph{Zero-shot prompts} contained only the query equation formatted under one of the five variants. \emph{N-shot prompts} include solved examples (up to ten) before the query, allowing us to probe in-context learning abilities of LLMs.\\  

\noindent
We use the same procedure to generate the data for both addition (\texttt{+}) and subtraction (\texttt{-}) operators. This procedure yields a balanced dataset, and systematically varied across operators, operand order, and prompt phrasings. We provide full dataset statistics in Table \ref{tab:dataset_stats}.

\subsection{LLM Inference}
For pretrained LLMs, we use greedy sampling with a temperature of 0 and sample up to 20 new tokens. For instruction-tuned LLMs, we use the sampling strategy and parameters suggested by the LLM creators. Additionally, we use the default system prompt for each instruction-tuned LLM and sample up to 500 new tokens.

We use a fixed seed across all experiments to ensure reproducibility. We extract the final numerical answer from the LLMs' generated text using a robust parsing mechanism. We run all experiments on 4x H100 GPUs using \texttt{vLLM}~\cite{kwon2023efficient} without any quantization. 

\section{Results}
In this section, we analyze LLM performance on subtraction, using addition as a reference point. We examine how accuracy varies across LLMs, operand order, and analyze the errors that emerge.

\subsection{Can LLMs subtract numbers?}
\label{subsec:addvssub}
\begin{figure}[ht]
  \centering
  \includegraphics[width=\columnwidth]{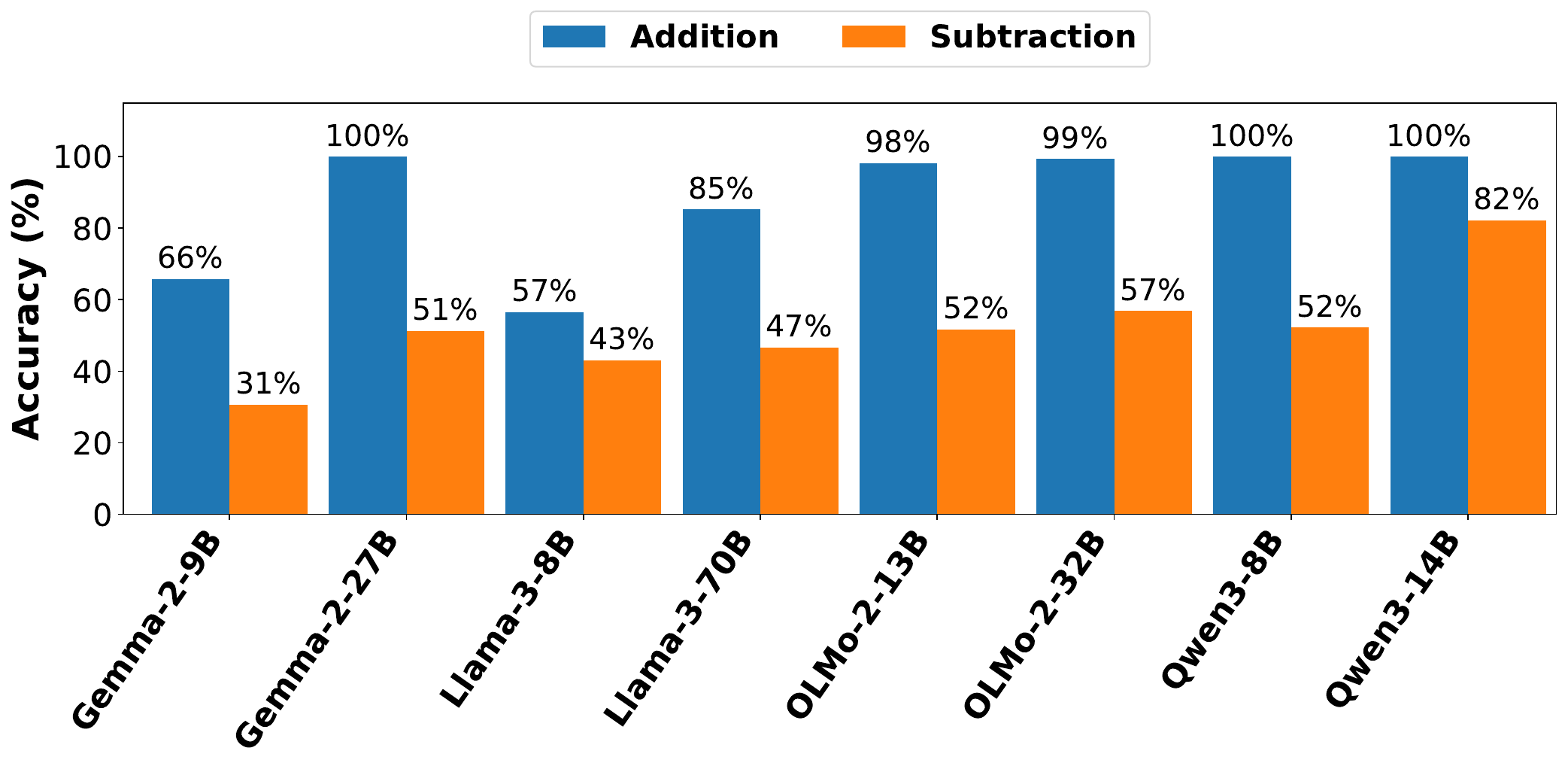}
  \caption{Zero-shot performance of LLMs on addition and subtraction problems (averaged across prompt variants). Subtraction is consistently harder, even for LLMs that perform well on addition.}
  \label{fig:q_1}
\end{figure}

We start our exploration by comparing subtraction accuracy with that of addition across prompt variants in a zero-shot setting. 
Figure~\ref{fig:q_1} shows that while most LLMs achieve near-perfect accuracy on addition, subtraction is substantially harder. For example, \texttt{Qwen3-8B} reaches almost 100\% on addition but only around 57\% on subtraction. Similarly, \texttt{OLMo-2-32B} scores above 99\% on addition, but drops to roughly 56\% on subtraction. In contrast, smaller LLMs such as \texttt{Gemma-2-9B} and \texttt{Llama-3-8B} remain poor at both tasks. In general, the performance of the subtraction lags the addition by 30 -- 50 points in several LLMs, showing that the LLMs struggle to compute the correct answer for the subtraction. This leads us to explore further if there are any patterns in the incorrect answers.

\subsection{Subtraction with \hmath $a>b$ versus \hmath $a<b$}
\label{subsec:posvsneg}

\begin{figure}[htpb]
  \centering
  \includegraphics[width=\columnwidth]{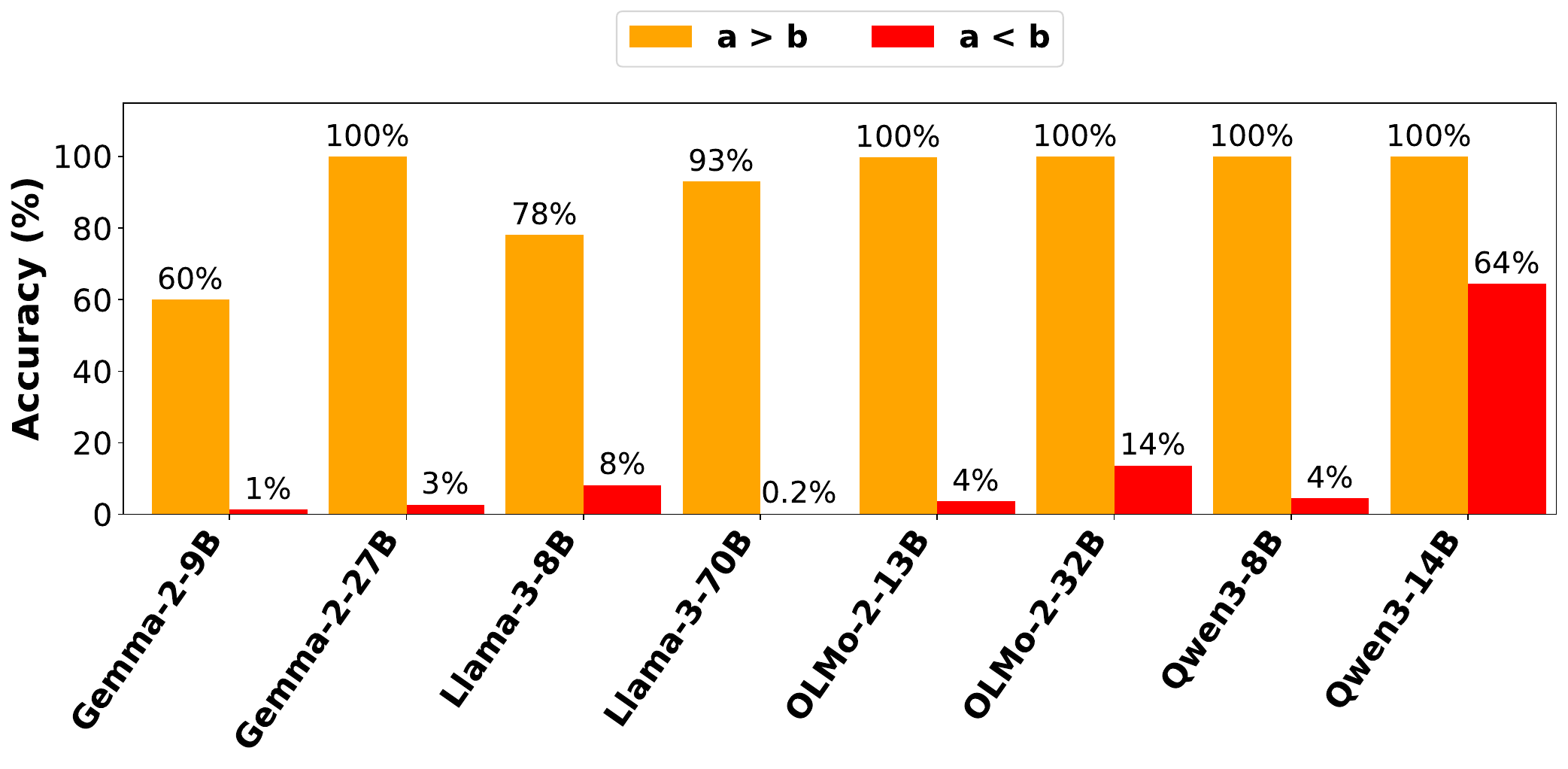}
  \caption{Zero-shot performance on subtraction ($a-b$) across operand and prompt variants. Pretrained LLMs perform subtraction well when $a>b$ but fail almost completely when $a<b$, showing a strong asymmetry.
  }
  \label{fig:q_2}
\end{figure}

We divide our input data into two subsets, $a > b$ and $a < b$, to see if there is any concrete error pattern exists there. Figure~\ref{fig:q_2} suggests a strong asymmetry in performance here as nearly all LLMs succeed in performing $a - b$ when $a > b$, but accuracy collapses for $a < b$.
For instance, \texttt{Qwen3-8B}, \texttt{Gemma-2-27B}, and \texttt{Llama-3.1-70B} achieve near-perfect scores when the answer is positive, but below 5\% when it is negative.

\begin{figure}[htpb]
  \centering
  \includegraphics[width=\columnwidth]{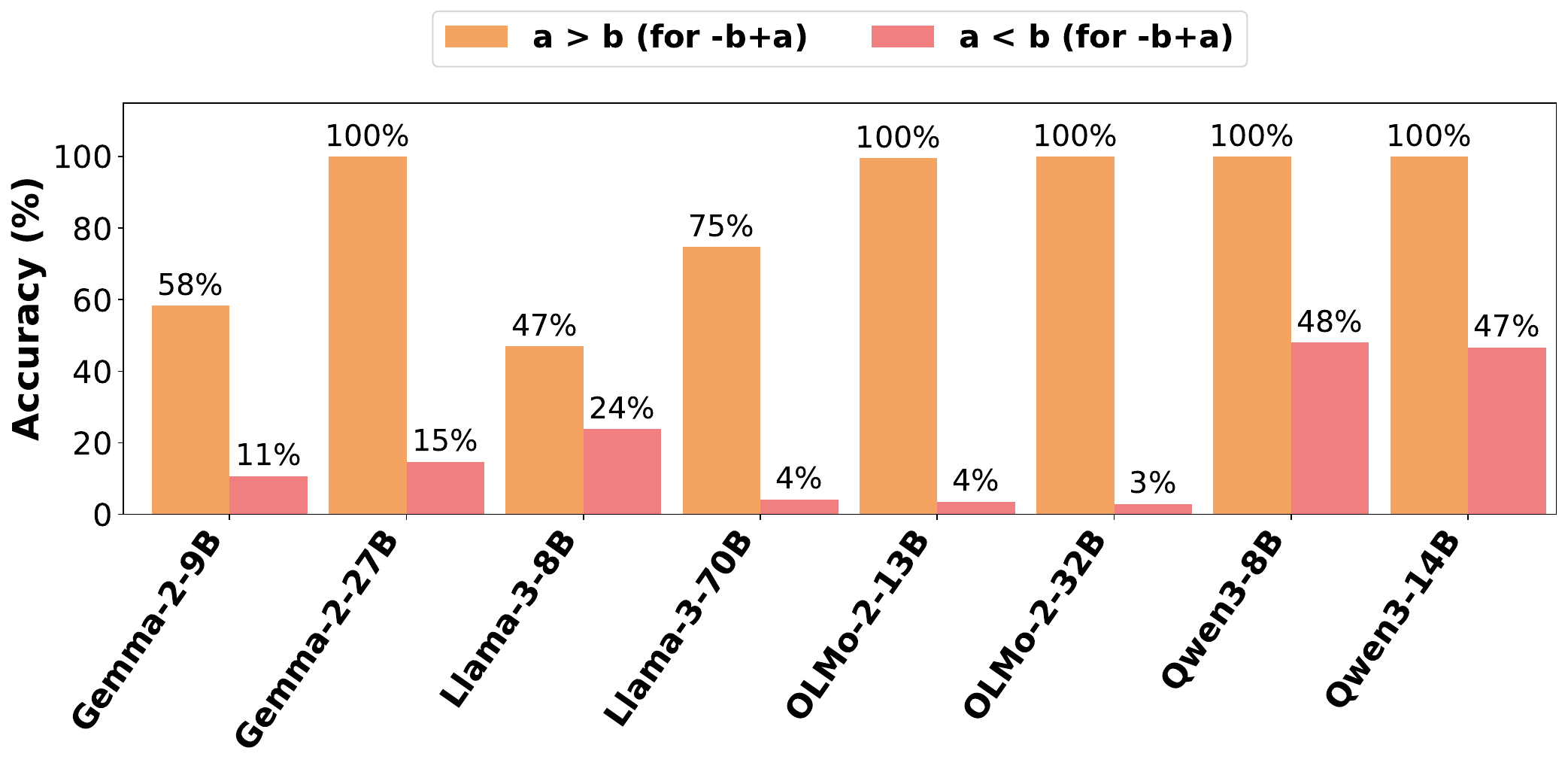} 
  \caption{Zero-shot performance on the \texttt{-b+a} input pairs. This plot shows the same asymmetry as Figure~\ref{fig:q_2}.}
  \label{fig:q_6}
\end{figure}
To confirm if this asymmetry is specific to the \texttt{a-b} template or a general difficulty with negative-signed results, we analyze the \texttt{-b+a} input type. We use the same original operands $a$ and $b$ to generate this data. We find an identical trend for \texttt{-b+a} as shown in Figure~\ref{fig:q_6}. This confirms that the models' failure is not due to the subtraction operation itself, but rather a systemic difficulty in producing a negative integer as the final answer.


\subsection{What errors do LLMs make when \hmath $a<b$?}
\label{subsec:absolutevals}

\begin{figure}[htpb]
  \centering
  \begin{subfigure}[b]{\columnwidth}
    \centering
    \captionsetup{labelformat=empty}
    \includegraphics[width=\columnwidth]{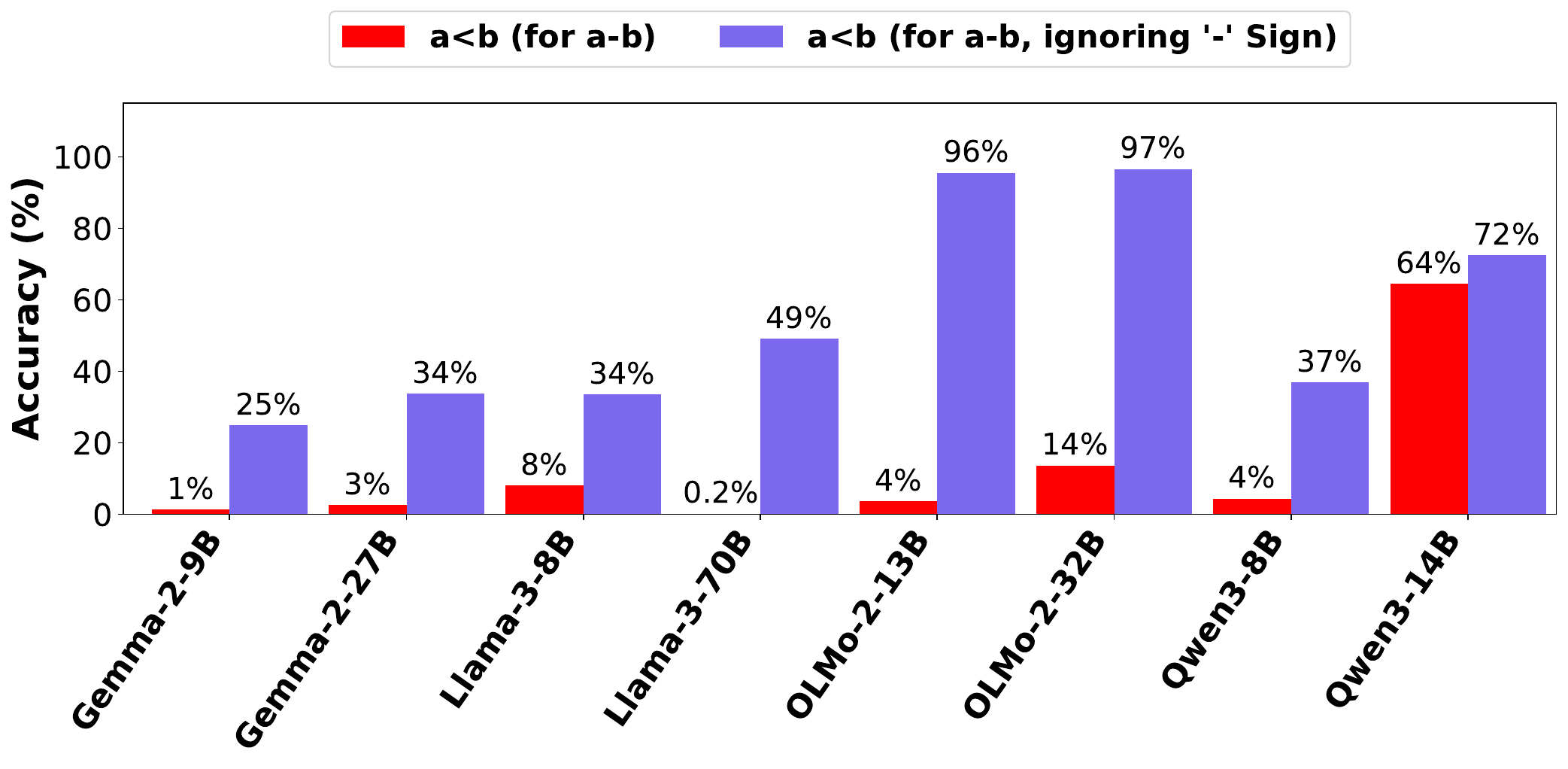}    
    \caption{}
    \label{fig:q_3_sub}
    
  \end{subfigure}
  \begin{subfigure}[b]{\columnwidth}
    \centering
    \captionsetup{labelformat=empty}
    \includegraphics[width=\columnwidth]{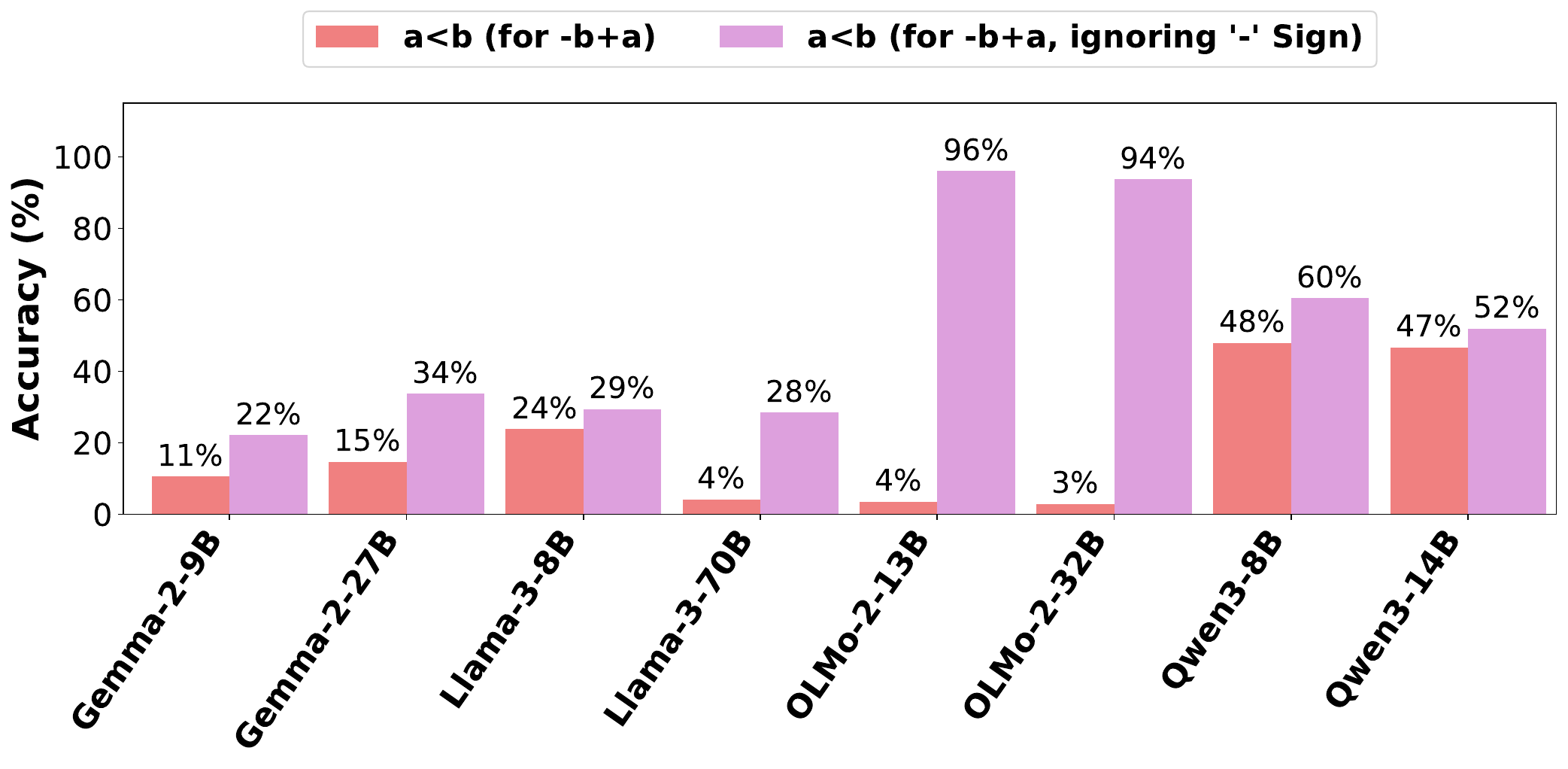}
    \caption{}
    \label{fig:q_6_sign_sub}
  \end{subfigure}
  \vspace{-10mm}
  \caption{Zero-shot performance on \texttt{a-b} (above) and \texttt{-b+a} (below). The gap between the accuracy with and without the `-' shows that pretrained LLMs often compute the correct magnitude but omit the negative sign.}
  \label{fig:q_3_combined}
\end{figure}

To isolate the source of errors on negative-result tasks, we measure the numerical accuracy without the negative sign. Figure~\ref{fig:q_3_combined} shows that accuracy jumps significantly under this relaxed metric for both the \texttt{a-b} and \texttt{-b+a} input pairs for all the LLMs.

This common error pattern indicates that LLMs may compute the correct magnitude but systematically omit the negative sign.

\subsection{Did LLMs simply \textit{forget} to include the sign?}

\label{subsec:probing}
The increase in accuracy when ignoring the negative sign (Figure~\ref{fig:q_3_combined}) raises the question of whether LLMs internally represent the sign even if they fail to output it.
To test this, we trained linear probes on the final layer activations of three representative LLMs: \texttt{Gemma-2 9B}, \texttt{Llama-3.1-8B}, and \texttt{Qwen3-8B}.
The probing task was deliberately simple: predict whether the result of a subtraction should be positive or negative.

The probes achieved near-perfect accuracy across all settings. For instance, 100\% on \texttt{Gemma-2 9B} and \texttt{Qwen3-8B}, and above 99\% on \texttt{Llama-3.1-8B}.
These numbers are averaged over five independent runs, with standard deviations below 0.1, confirming the stability of the results.

Consequently, while LLMs often omit the negative sign in their textual outputs, their hidden states consistently distinguish between positive and negative outcomes. This indicates a disconnect between representation and generation: LLMs reflect internally that there is a difference between positive and negative results, but this knowledge is not faithfully transferred to the decoding stage.

For all the results shown in Section \ref{subsec:addvssub}, \ref{subsec:posvsneg}, \ref{subsec:absolutevals}, and \ref{subsec:probing}, we observe the same trend for the multi-token input numbers. We provide the results for the same in Appendix \ref{sec:oov_apdx}.

\section{Can we make LLMs better for \hmath$a<b$?}
\label{sec:improvements}

\subsection{In-context learning with pretrained LLMs}
\label{sub:pt_in_ctx}

Literature suggests that LLMs can benefit from in-context examples~\cite{brown2020fewshot,agarwal2024manyshot}. Therefore, we evaluate pretrained LLMs with 3, 5, and 10-shot in-context examples to assess whether this improves subtraction performance for the case of $a<b$. 

As shown in Table~\ref{tab:q_5a}, few-shot prompting leads to mixed results. Some LLMs exhibit moderate gains, but the improvements are neither uniform nor consistent across families or sizes. For instance, \texttt{Llama-3.1-8B} improves substantially from 8.1\% to 31.5\% accuracy (with the negative sign) when moving from zero- to five-shot prompting, suggesting limited but non-trivial in-context learning. \texttt{Qwen3-14B} also benefits initially, though its performance plateaus beyond three shots. Conversely, larger LLMs such as \texttt{Llama-3.1-70B} and \texttt{Gemma-2-27B} remain unstable, showing marginal or inconsistent gains with additional examples. 

Across nearly all LLMs, accuracy is considerably higher when the negative sign is ignored, often exceeding 90\%. This indicates that LLMs frequently compute the correct magnitude but omit the sign, consistent with the analysis in Section~\ref{sec:improvements}. 

In summary, while in-context examples can occasionally help pretrained LLMs handle negative results, the overall effect is modest and inconsistent. Complete results for all shot settings (3, 5, and 10) are provided in Appendix~\ref{sec:few_shot_full_apdx}.

\begin{table}[htpb]
\centering
\resizebox{\linewidth}{!}{
\begin{tabular}{lrrrr}
\toprule
LLM & \multicolumn{2}{c}{0-shot} & \multicolumn{2}{c}{5-shot} \\
 & w (-) & w/o (-) & w (-) & w/o (-) \\
\midrule
Gemma-2-9B  &          1.33 &        24.89 &          0.00 &        95.56 \\
Gemma-2-27B &          2.67 &        33.78 &          0.00 &        92.00 \\
Llama-3-8B  &          8.13 &        33.51 &         31.46 &        87.72 \\
Llama-3-70B &          0.22 &        49.11 &          1.52 &        90.71 \\
OLMo-2-13B  &          3.70 &        95.55 &         18.97 &        99.09 \\
OLMo-2-32B  &         13.65 &        96.55 &          5.09 &        94.51 \\
Qwen3-8B    &          4.44 &        36.89 &          4.89 &        96.44 \\
Qwen3-14B   &         64.44 &        72.44 &         55.11 &        90.67 \\
\bottomrule
\end{tabular}
}
\caption{Few-shot prompting accuracy (\%) of pretrained LLMs with (-) sign and without the (-) sign.}
\label{tab:q_5a}
\end{table}

\subsection{Instruction-tuned LLMs}
\label{sub:it_llm}
The instruction-tuned LLMs report strong performance on various math benchmarks such as \texttt{MATH} and \texttt{GSM8k}. Consequently, we test the instruction-tuned variants of our pretrained LLMs on this subtraction subtask. Our results suggest that almost all instruction-tuned LLMs reach above 90\% accuracy (Table~\ref{tab:q_5b}), including LLMs that fail completely in their pretrained variants.


\begin{table}[htbp]
\centering
\resizebox{\linewidth}{!}{%
\begin{tabular}{lrr}
\toprule
LLM & \shortstack{Pretrained} & \shortstack{Instruction-Tuned} \\
\midrule
Gemma-2-9B & 1.33 & 100.00 \\
Gemma-2-27B & 2.67 & 100.00 \\
Llama-3-8B & 8.13 & 91.42 \\
Llama-3-70B & 0.22 & 99.91 \\
OLMo-2-13B & 3.70 & 99.54 \\
OLMo-2-32B & 13.65 & 88.27 \\
Qwen3-8B & 4.44 & 100.00 \\
Qwen3-14B & 64.44 & 100.00 \\
\bottomrule
\end{tabular}
}
\caption{Zero-shot accuracy (\%) of pretrained and instruction-tuned LLMs. 
Instruction tuning substantially improves performance on arithmetic queries, 
often pushing accuracy above 90\%.}
\label{tab:q_5b}
\end{table}

We speculate that these gains come from the instruction fine-tuning stage.
Although it is not possible to investigate the instruction-tuning data for all the LLMs, it is possible for \texttt{OLMo-2}. We investigate \texttt{OLMo-2} instruction-tuning dataset and find that it contains \texttt{MATH}~\cite{hendrycks2021math} training set, \texttt{GSM8k}~\cite{cobbe2021gsm8k} training set and \texttt{Tülu 3}~\cite{lambert2024tulu} dataset. All these datasets contain subtraction data including input cases where $a<b$. For the same, we speculate that \texttt{OLMo-2} variants benefit from this additional data seen during instruction fine-tuning (for exact fine-tuning prompts refer to Appendix~\ref{sec:olmo_apdx}).

\section{Conclusion}
Taken together, our results highlight subtraction as a persistent challenge for current LLMs. Even LLMs that solve addition near perfectly often falter when subtraction yields a negative result, typically omitting the sign. Probing experiments suggest that the relevant information is present in hidden states but does not consistently transfer to the output layer, underscoring a mismatch between representation and generation. Few-shot prompting reduces errors for most pretrained LLMs but remains unstable, while instruction-tuned LLMs consistently perform better in most cases and achieve near-perfect performance. Thus, subtraction can provide a valuable diagnostic lens on how LLMs handle basic arithmetic. For the same, subtraction should be given equal importance as a standard testbed for evaluating numerical reasoning in future LLM research.

\section*{Limitations}

While we test the LLMs on a reasonably large test, we only ran inference using three seeds. We also could not test any closed-source LLMs since their pretrained versions are not available via APIs. 

\section*{Acknowledgement}
This research is funded in part by the Deutsche Forschungsgemeinschaft (DFG, German Research Foundation) -- Project-ID 471607914 -- GRK 2853/1 ``Neuroexplicit Models of Language, Vision, and Action''; and Project-ID 389792660 -- TRR~248 ``Foundations of Perspicuous Software Systems''. The authors gratefully acknowledge the computing time granted by the John von Neumann Institute for Computing (NIC) and provided on the supercomputer JURECA at Jülich Supercomputing Centre (JSC). The authors gratefully acknowledge the computing time made available to them on the high-performance computer at the NHR Center of TU Dresden. This center is jointly supported by the Federal Ministry of Research, Technology and Space of Germany and the state governments participating in the NHR (www.nhr-verein.de/unsere-partner). The authors would like to thank Alisa Kovtunova, Aleksandra Bakalova, Iza Škrjanec, Sukrut Rao, Yash Sarrof for discussions and feedback on the draft.



\bibliography{custom}

\appendix

\section{Multi-token performance}
\label{sec:oov_apdx}
We run the inference on LLMs and analyze their performance using numbers that were split into multiple tokens as well. We test up to 3 tokens for a single number. For generating multi-token data, we use the same procedure described in Section \ref{subsec:data}. Here, the operand range is between the maximum number in the tokenizer vocab + 1 to the maximum number in the tokenizer vocab $\times$ \texttt{100}. We describe our findings for these experiments in the following subsections.

\subsection{Can LLMs subtract multi-token numbers?}
We observe an expected drop in performance for multi-token addition predicted by \cite{yang2025number}, but the trend of lower subtraction performance across all LLMs still holds for the multi-token subset as well. We provide the results in Figure \ref{fig:q_1_oov} below.

\begin{figure}[h]
  \centering
  \includegraphics[width=\columnwidth]{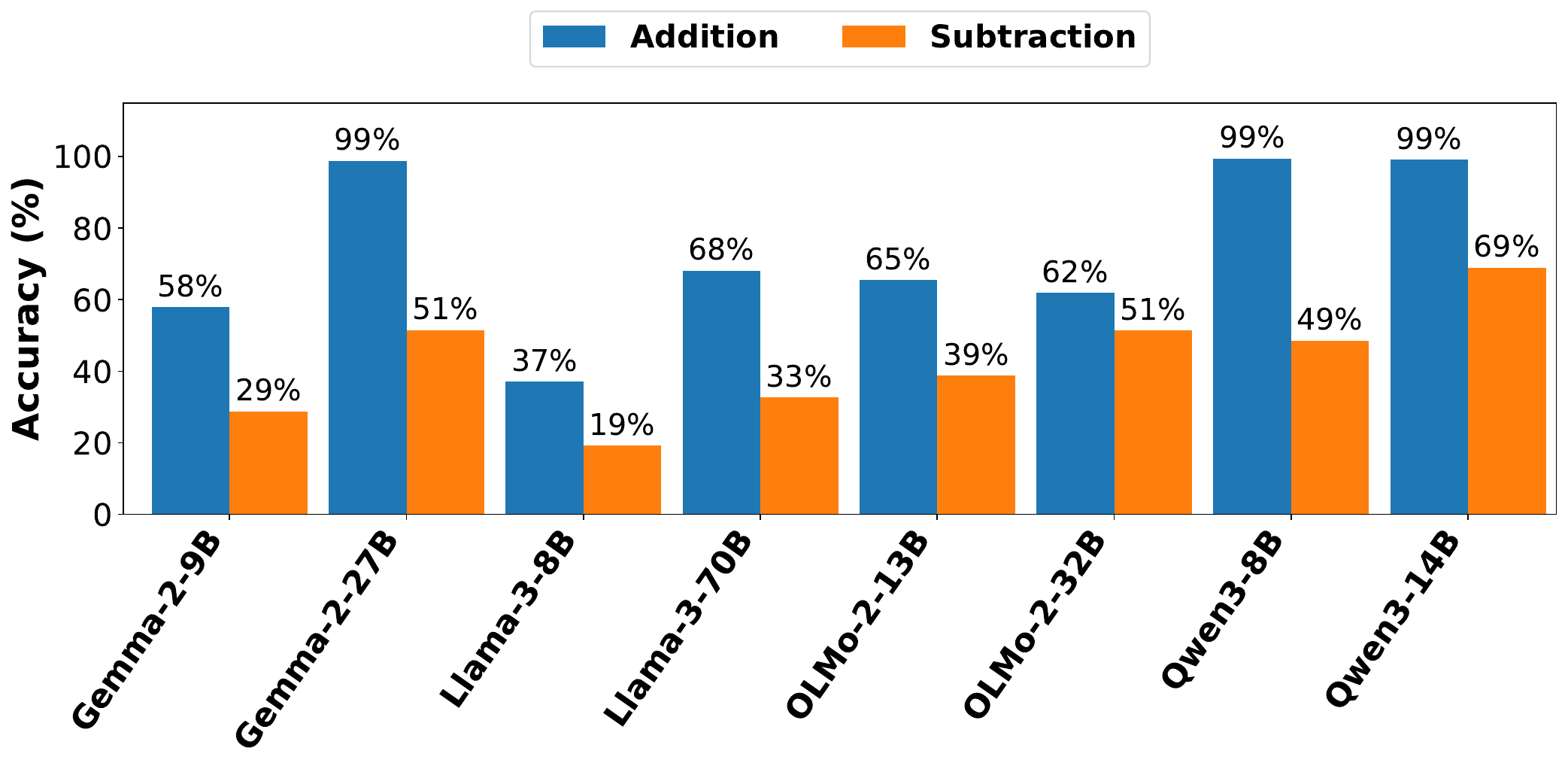}
  \caption{Zero-shot performance on multi-token addition and subtraction problems (averaged across prompt variants). Similar to single-token subtraction, multi-token subtraction is also consistently harder for all the LLMs.}
  \label{fig:q_1_oov}
\end{figure}

\subsection{Multi-token Subtraction with \hmath $a>b$ versus \hmath $a<b$}
\begin{figure}[htpb]
  \centering
  \includegraphics[width=\columnwidth]{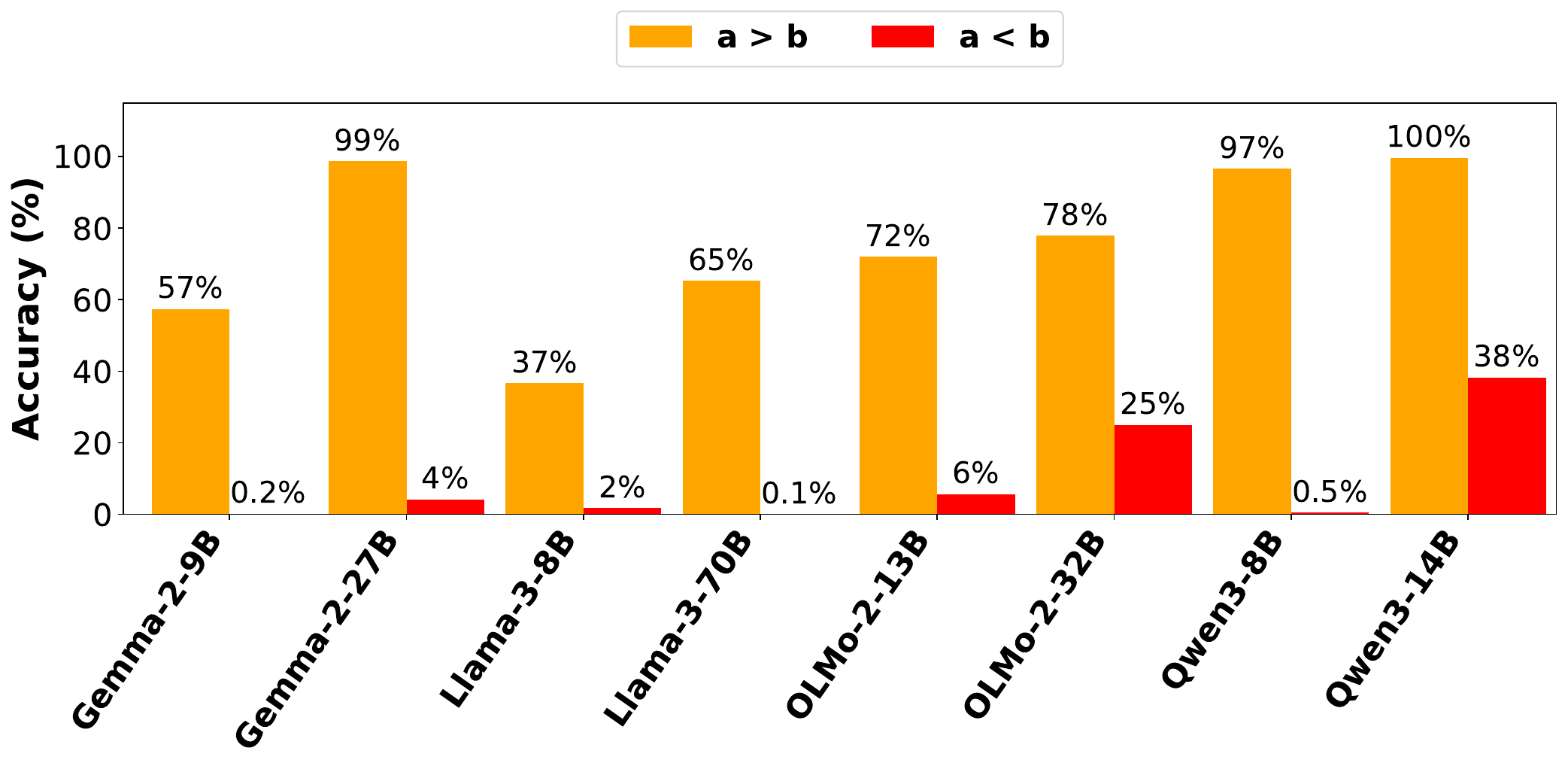}
  \caption{Zero-shot performance on Multi-token subtraction across operand and prompt variants. Pretrained LLMs do not perform multi-token subtraction as well as the single-token subtraction, but the $a>b$ vs $a<b$ performance asymmetry still holds.}
  \label{fig:q_2_oov}
\end{figure}

The trend of lower performance on the $a<b$ subset compared to the $a>b$ subset across all LLMs holds for the multi-token subset as well. We provide these results in Figure \ref{fig:q_2_oov}. 

\subsection{What goes wrong for \hmath $a<b$ in the multi-token subset?}
\begin{figure}[h]
  \centering
  \includegraphics[width=\columnwidth]{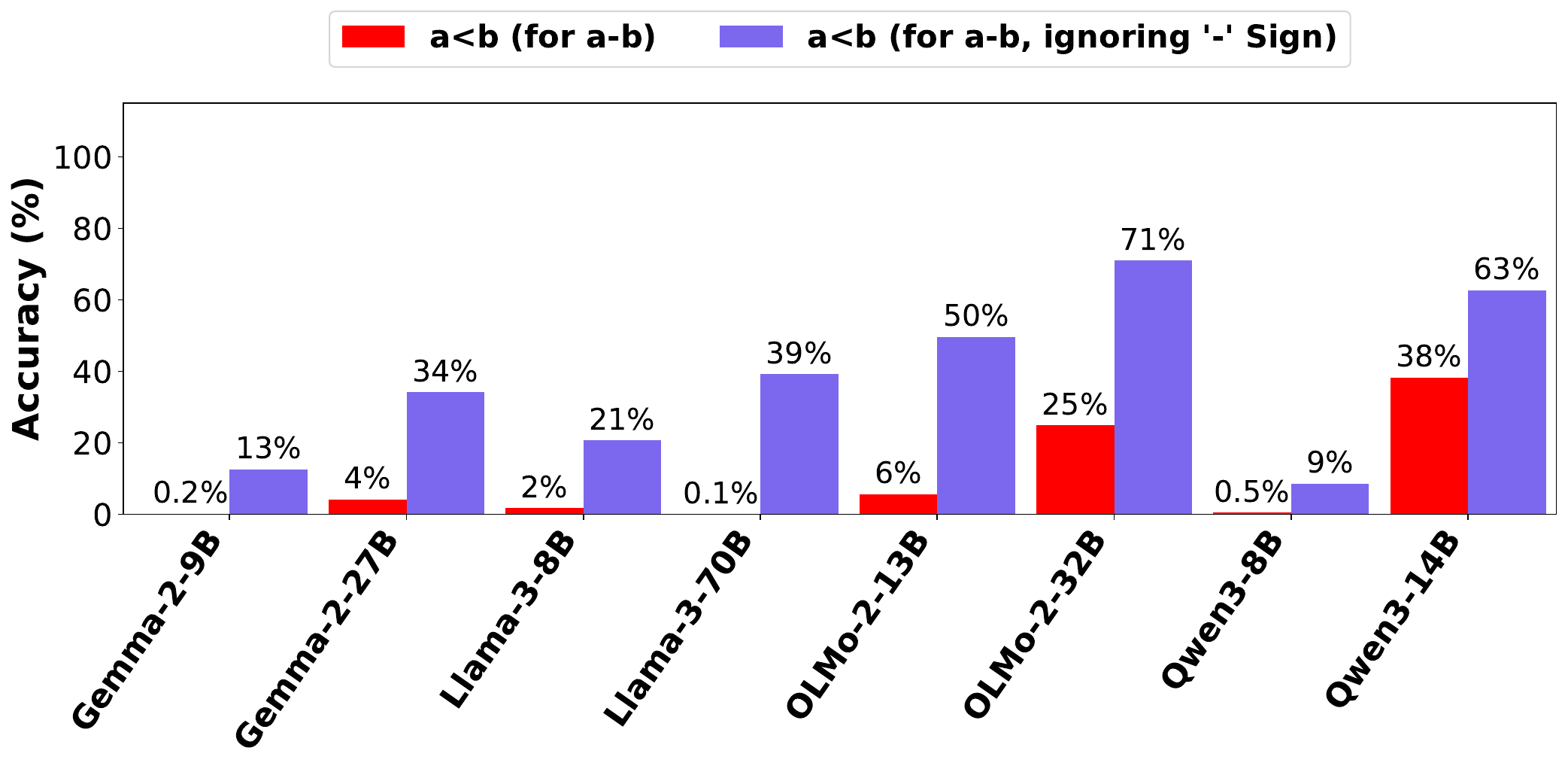}
  \caption{Zero-shot performance on $a<b$ examples for the Multi-token subset. The consistent gap between the accuracy and the accuracy ignoring sign across all the LLMs shows that pretrained LLMs often compute the correct magnitude but fail to output the negative sign.}
  \label{fig:q_3_oov}
\end{figure}

The trend of missing `--' sign on the $a<b$ subset across all LLMs holds for the multi-token subset as well. We provide these results in Figure~\ref{fig:q_3_oov}.

\section{Few-shot additional results}
\label{sec:few_shot_full_apdx}

\begin{table*}[htpb]
\centering
\begin{tabular}{lrrrrrrrr}
\toprule
LLM & \multicolumn{2}{c}{0-shot} & \multicolumn{2}{c}{3-shot} & \multicolumn{2}{c}{5-shot} & \multicolumn{2}{c}{10-shot} \\
 & w (-) & w/o (-) & w (-) & w/o (-) & w (-) & w/o (-) & w (-) & w/o (-) \\
\midrule
Gemma-2-9B  &          1.33 &        24.89 &          0.00 &        92.00 &          0.00 &        95.56 &          0.00 &        99.56 \\
Gemma-2-27B &          2.67 &        33.78 &          0.44 &        72.89 &          0.00 &        92.00 &          0.00 &        93.33 \\
Llama-3-8B  &          8.13 &        33.51 &         21.94 &        85.14 &         31.46 &        87.72 &         29.28 &        86.08 \\
Llama-3-70B &          0.22 &        49.11 &          1.49 &        71.82 &          1.52 &        90.71 &          0.77 &        97.49 \\
OLMo-2-13B  &          3.70 &        95.55 &         19.92 &        98.63 &         18.97 &        99.09 &         18.27 &        99.01 \\
OLMo-2-32B  &         13.65 &        96.55 &          5.21 &        93.96 &          5.09 &        94.51 &          3.89 &        82.61 \\
Qwen3-8B    &          4.44 &        36.89 &          4.44 &        93.78 &          4.89 &        96.44 &          2.22 &        92.89 \\
Qwen3-14B   &         64.44 &        72.44 &         61.78 &        80.44 &         55.11 &        90.67 &         15.11 &        88.89 \\
\bottomrule
\end{tabular}
\caption{Few-shot prompting accuracy (\%) of pretrained LLMs with (-) sign and without the (-) sign}
\label{tab:all_shot_comparison}
\end{table*}

\section{OLMo Instruction-tuning prompt}
\label{sec:olmo_apdx}
The prompt used in the Tulu dataset for fine-tuning \texttt{OLMo} models is given in Figure \ref{fig:olmo_prompt}. 

\begin{figure*}[htpb]
\centering
\begin{tcolorbox}[title={Math Problems (response)}, width=\textwidth]
\textbf{Provide solution to the given math problem.}

\medskip
\textbf{Problem:} \texttt{\{math\_problem\}}

\medskip
\textbf{Note:} Provide your solution step-by-step, and end your solution in a new line in the following format: \\[2mm]
\textbf{Final Answer:} The final answer is $final\_answer$.
\end{tcolorbox}
\caption{Prompt used for math word problems.}
\label{fig:olmo_prompt}
\end{figure*}

\section{Prompt details}
\label{sec:prompt_apdx}
Here, we provide all the prompts used in our experiments verbatim. Figure \ref{fig:prompt_formats} contains the zero-shot variants. 
\lstset{
  basicstyle=\ttfamily\footnotesize,
  columns=fullflexible,
  keepspaces=true,
  aboveskip=2pt,
  belowskip=2pt,
  frame=none,
  xleftmargin=0pt
}

\begin{figure*}[htpb]
\centering
\begin{minipage}{\linewidth}
\vspace{2pt}

\begin{enumerate}[leftmargin=*, itemsep=2pt, topsep=2pt]
  \item \begin{lstlisting}
### a {op} b = \boxed{{
\end{lstlisting}

  \item \begin{lstlisting}
Solve the problem below and give only the simplified answer in \boxed{answer}.
### Problem: a {op} b =
Answer: \boxed{
\end{lstlisting}

  \item \begin{lstlisting}
Solve the problem by computing the {op}, and report only the final answer in \boxed{answer}.
### Problem: a {op} b =
Answer: \boxed{
\end{lstlisting}

  \item \begin{lstlisting}
Find the {op} ({op_sign}) for the problem below and respond only with \boxed{answer}.
### Problem: a {op} b =
Answer: \boxed{
\end{lstlisting}

  \item \begin{lstlisting}
Perform the {op} ({op_sign}) to solve the problem; output only \boxed{answer}.
### Problem: a {op} b =
Answer: \boxed{
\end{lstlisting}
\end{enumerate}
\end{minipage}
\caption{Zero-shot prompt variants used in our experiments. Here, $a$ and $b$ are numbers; \texttt{op} is either \emph{sum} or \emph{difference}; \texttt{op\_sign} is $+$ or $-$. Each prompt requires the LLMs to return the answer inside a boxed expression.}
\label{fig:prompt_formats}
\end{figure*}

\section{Negative sign probing}
\label{sec:probing_apdx}
Table \ref{tab:appendix-probe-details} contains all the details for each run.
\begin{table*}[htpb]
    \centering
    
    \begin{tabular}{llcccccc}
        \toprule
        \textbf{LLM} & \textbf{Probe Type} & \textbf{Run 1} & \textbf{Run 2} & \textbf{Run 3} & \textbf{Run 4} & \textbf{Run 5} & \textbf{Mean $\pm$ Std. Dev.} \\
        \midrule
        \multirow{3}{*}{Gemma-2 9B} 
        & Single-token & $100.00$ & $100.00$ & $100.00$ & $100.00$ & $100.00$ & $100.00 \pm 0.00$ \\
        & Multi-token & $99.52$ & $99.72$ & $99.60$ & $99.56$ & $99.60$ & $99.60 \pm 0.07$ \\
        & Combined & $99.82$ & $99.70$ & $99.82$ & $99.86$ & $99.66$ & $99.77 \pm 0.08$ \\
        \midrule
        \multirow{3}{*}{Llama-3.1 8B} 
        & Single-token & $99.60$ & $99.48$ & $99.72$ & $99.88$ & $99.64$ & $99.66 \pm 0.13$ \\
        & Multi-token & $99.00$ & $98.88$ & $99.20$ & $99.24$ & $99.16$ & $99.10 \pm 0.14$ \\
        & Combined & $99.26$ & $99.46$ & $99.30$ & $99.32$ & $99.50$ & $99.37 \pm 0.09$ \\
        \midrule
        \multirow{3}{*}{Qwen3 8B} 
        & Single-token & $100.00$ & $100.00$ & $100.00$ & $100.00$ & $100.00$ & $100.00 \pm 0.00$ \\
        & Multi-token & $99.92$ & $99.88$ & $99.92$ & $100.00$ & $100.00$ & $99.94 \pm 0.05$ \\
        & Combined & $99.83$ & $100.00$ & $100.00$ & $99.93$ & $100.00$ & $99.95 \pm 0.07$ \\
        \bottomrule
    \end{tabular}
    \caption{Detailed probe accuracy results across all LLMs and runs. All values are percentages (\%).}
    \label{tab:appendix-probe-details}
\end{table*}

\end{document}